%% file: main.tex
\documentclass[11pt,a4paper]{article}
\usepackage[hyperref]{acl2020}
\usepackage{times}
\usepackage{latexsym}
\usepackage{bbding}
\usepackage{listings}

\usepackage{relsize}

\usepackage{microtype}

\usepackage{amsmath, amssymb, amsthm}
\usepackage{bbm}
\usepackage{booktabs}
\usepackage{graphicx}
\usepackage{makecell}
\usepackage{hhline}

\usepackage{multirow}

\usepackage{xcolor}
\usepackage{CJKutf8}

\usepackage{pgfplotstable}
\usepackage{pgfplots}

\usepackage{url}
\usepackage[normalem]{ulem}

\usepackage[T5,T1]{fontenc}
\usepackage[utf8]{inputenc}

\newcommand\stanzafamily[1]{{\usefont{T1}{quicksand}{m}{n} #1}}

\DeclareTextSymbolDefault{\OHORN}{T5}
\DeclareTextSymbolDefault{\UHORN}{T5}
\DeclareTextSymbolDefault{\ohorn}{T5}
\DeclareTextSymbolDefault{\uhorn}{T5}

\input{std-macros.tex}

\newcommand{\fone}{$\T{F}_1$}
\newcommand\ph[1]{\phantom{#1}}

\definecolor{lblue}{HTML}{A6CEE3}
\definecolor{lgreen}{HTML}{B2DF8A}
\definecolor{lred}{HTML}{FB9A99}
\definecolor{lorange}{HTML}{FDBF6F}
\definecolor{mblue}{HTML}{80B1D3}
\definecolor{mgreen}{HTML}{B3DE69}
\definecolor{mred}{HTML}{FB8072}
\definecolor{morange}{HTML}{FDB462}
\definecolor{blue}{HTML}{1F78B4}
\definecolor{green}{HTML}{33A02C}
\definecolor{red}{HTML}{E31A1C}
\definecolor{orange}{HTML}{FF7F00}
\definecolor{dblue}{HTML}{08519C}
\definecolor{dgreen}{HTML}{006D2C}
\definecolor{dorange}{HTML}{EC7014}

\newcommand{\stanzatextmark}{St\hspace{.1ex}a\hspace{.65ex}n\hspace{.35ex}z\hspace{.55ex}a\hspace{.7ex}}
\newcommand{\stanzatitle}{{\stanzafamily{\stanzatextmark}}}
\newcommand{\stanza}{{\stanzafamily{\relsize{-0.5}\stanzatextmark}}}
\newcommand{\corenlp}{CoreNLP}
\newcommand{\flair}{\textsc{Flair}}
\newcommand{\spacy}{spaCy}
\newcommand{\udpipe}{UDPipe}

\newcommand{\stanzalogo}{\raisebox{-.15\height}{\includegraphics[height=1.6\fontcharht\font`\B]{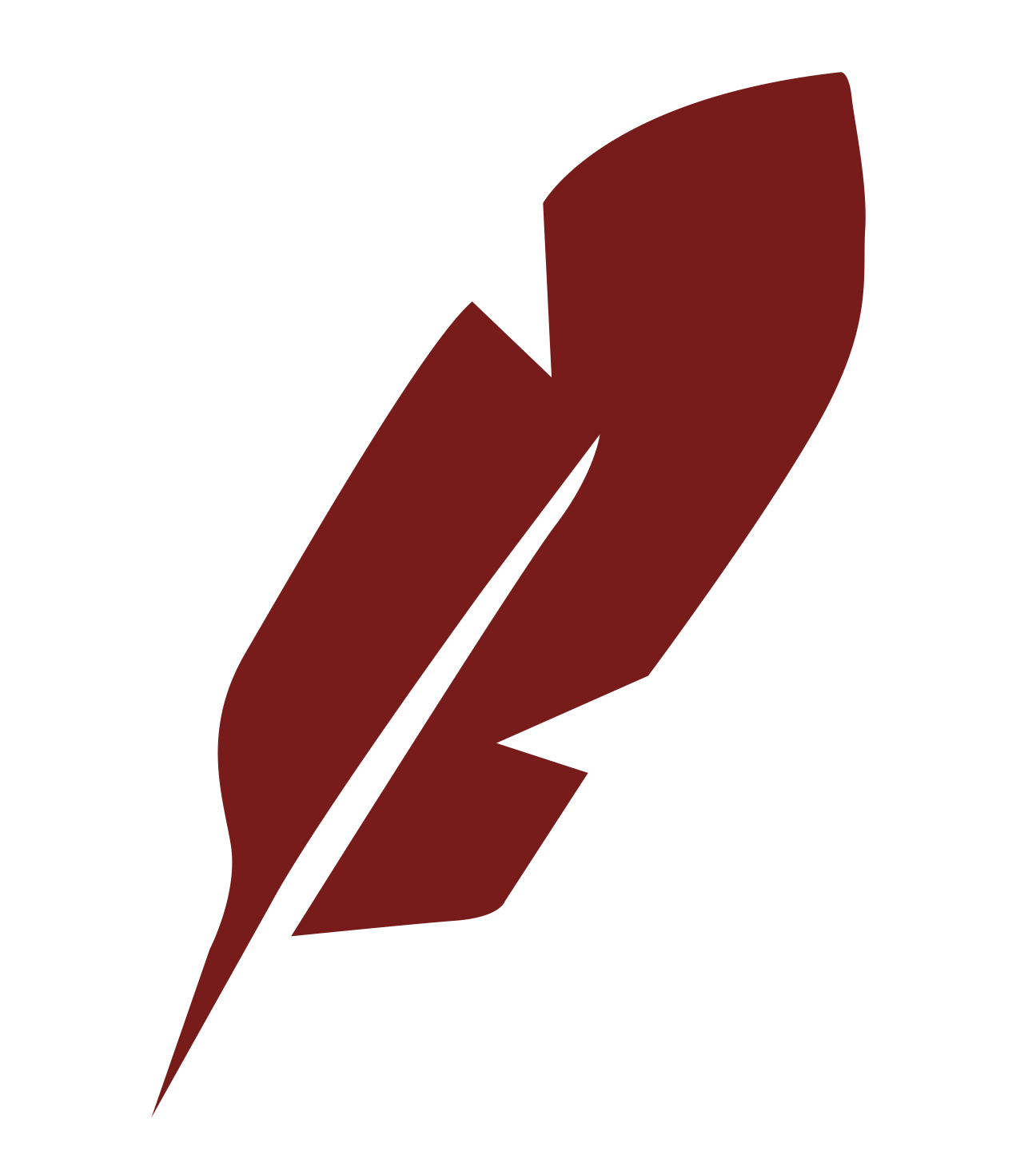}}}

\newcommand{\eg}{\textit{e.g.}}

\aclfinalcopy 

\newcommand{\nlang}{66}
\newcommand{\nnerlang}{8}
\newcommand{\ndataset}{112} 
\newcommand{\ntreebank}{100}
\newcommand{\nnerdataset}{12}

\title{\stanzalogo{}\stanzatitle{}: A Python Natural Language Processing Toolkit\\
for Many Human Languages}

\author{
Peng Qi*\quad Yuhao Zhang* \quad Yuhui Zhang\quad \\
\textbf{Jason Bolton\quad Christopher D. Manning} \\
Stanford University\\
Stanford, CA 94305\\
  {\tt \{pengqi, yuhaozhang, yuhuiz\}@stanford.edu}\\
  {\tt \{jebolton, manning\}@stanford.edu}\\
\\
}

\date{}

\begin{document}
\maketitle

\renewcommand{\thefootnote}{\fnsymbol{footnote}}
\footnotetext[1]{Equal contribution. Order decided by a tossed coin.}
\renewcommand{\thefootnote}{\arabic{footnote}}

\input{abstract}
\input{introduction}
\input{architecture}
\input{usage}
\input{experiments}
\input{conclusion}

\input{acknowledgments}

\bibliography{main}
\bibliographystyle{acl_natbib}


\end{document}

%% file: std-macros.tex
\newcommand\T{\text}

\newcommand\reffig[1]{Figure~\ref{fig:#1}}

\newcommand\reftab[1]{Table~\ref{tab:#1}}

\usepackage{color}



%% file: abstract.tex
\begin{abstract}
We introduce \stanza, an open-source Python natural language processing toolkit supporting \nlang{} human languages.
Compared to existing widely used toolkits, \stanza{} features a language-agnostic fully neural pipeline for text analysis, including tokenization, multi-word token expansion, lemmatization, part-of-speech and morphological feature tagging, dependency parsing, and named entity recognition.
We have trained \stanza{} on a total of \ndataset{} datasets, including the Universal Dependencies treebanks and other multilingual corpora, and show that the same neural architecture generalizes well and achieves competitive performance on all languages tested.
Additionally, \stanza{} includes a native Python interface to the widely used Java Stanford CoreNLP software, which further extends its functionality to cover other tasks such as coreference resolution and relation extraction.
Source code, documentation, and pretrained models for \nlang{} languages are available at \url{https://stanfordnlp.github.io/stanza/}.
\end{abstract}

%% file: introduction.tex
\section{Introduction}

The growing availability of open-source natural language processing (NLP) toolkits has made it easier for users to build tools with sophisticated linguistic processing.
While existing NLP toolkits such as \corenlp~\cite{manning-EtAl:2014:P14-5}, \flair~\cite{akbik-etal-2019-flair}, \spacy%
\footnote{\url{https://spacy.io/}}, and \udpipe~\cite{straka-2018-udpipe} have had wide usage, they also suffer from several limitations.
First, existing toolkits often support only a few major languages.
This has significantly limited the community's ability to process multilingual text.
Second, widely used tools are sometimes under-optimized for accuracy either due to a focus on efficiency (\eg, \spacy) or use of less powerful models (\eg, \corenlp), potentially misleading downstream applications and insights obtained from them.
Third, some tools assume input text has been tokenized or annotated with other tools, lacking the ability to process raw text within a unified framework. 
This has limited their wide applicability to text from diverse sources.

\begin{figure}
    \centering
    \includegraphics[width=0.48\textwidth]{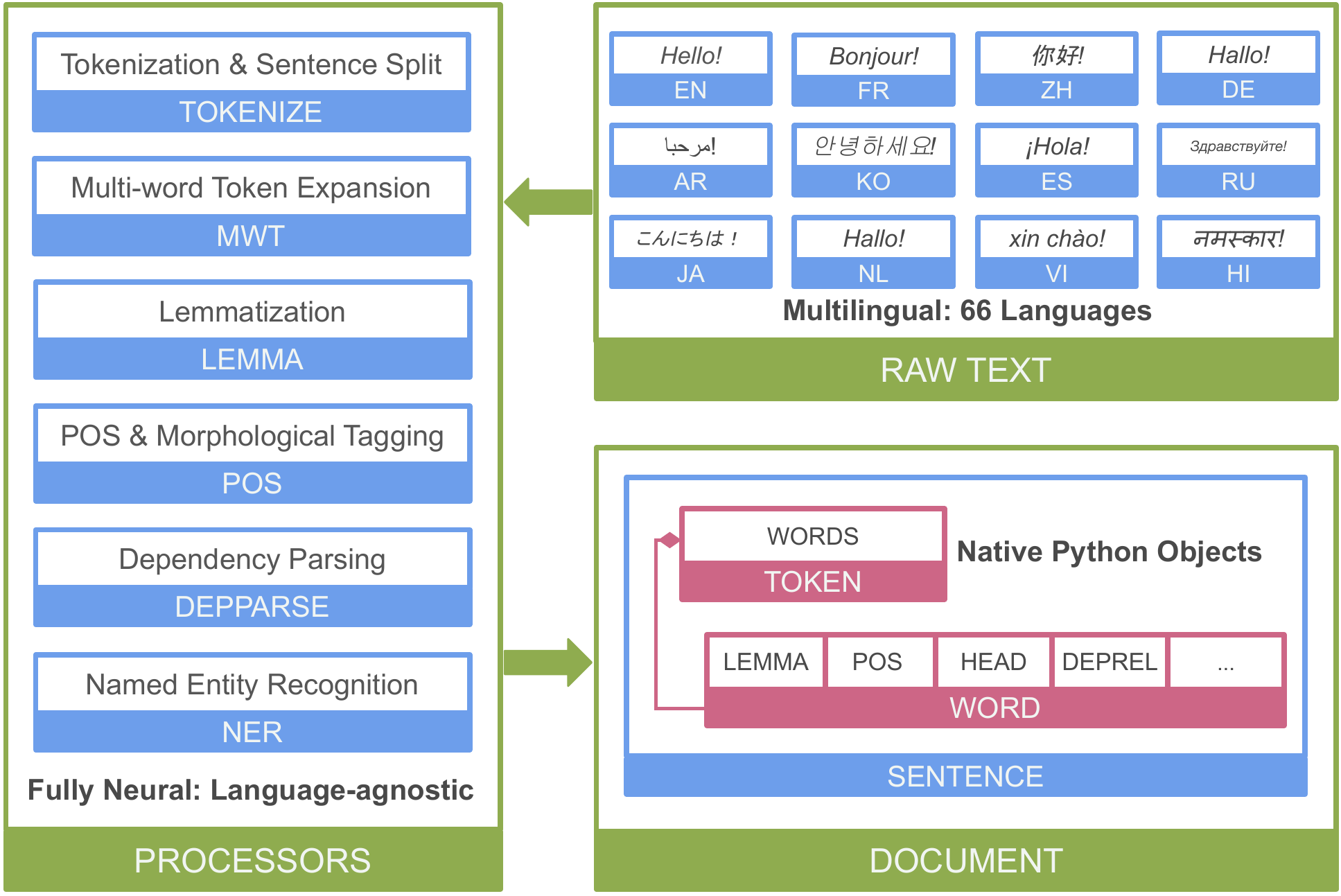}
    \caption{Overview of \stanza{}'s neural NLP pipeline.
    \stanza{} takes multilingual text as input, and produces annotations accessible as native Python objects.
    Besides this neural pipeline, \stanza{} also features a Python client interface to the Java CoreNLP software.}
    \label{fig:architecture}
\end{figure}

\begin{table*}
\small
    \centering
    \renewcommand{\arraystretch}{1.2}
    \newcolumntype{P}[1]{>{\centering\arraybackslash}p{#1}}
    \begin{tabular}{lP{1.8cm}P{1.8cm}P{1.8cm}P{1.5cm}P{1.8cm}P{2cm}P{1.8cm}}
    \toprule
        System & \# Human Languages & Programming Language & Raw Text Processing & Fully Neural & Pretrained Models & State-of-the-art Performance\\
    \midrule
        \corenlp & 6 & Java  & \Checkmark & & \Checkmark \\
        \flair & 12 & Python & & \Checkmark & \Checkmark & \Checkmark \\
        \spacy & 10 & Python & \Checkmark & & \Checkmark \\
        \udpipe & 61 & C++ & \Checkmark &  & \Checkmark & \Checkmark \\
    \midrule
        \stanzatitle{} & \nlang{} & Python & \Checkmark & \Checkmark & \Checkmark & \Checkmark \\ 
    \bottomrule
    \end{tabular}
    \caption{Feature comparisons of \stanza{} against other popular natural language processing toolkits.}
    \label{tab:feature_comparison}
\end{table*}

We introduce \stanza\footnote{The toolkit was called StanfordNLP prior to v1.0.0.}, a Python natural language processing toolkit supporting many human languages.
As shown in \reftab{feature_comparison}, compared to existing widely-used NLP toolkits, \stanza{} has the following advantages:
\vspace{-.3em}
\begin{itemize}
\itemsep0.2em
  \item \textbf{From raw text to annotations.} \stanza{} features a fully neural pipeline which takes raw text as input, and produces annotations including tokenization, multi-word token expansion, lemmatization, part-of-speech and morphological feature tagging, dependency parsing, and named entity recognition.
  \item \textbf{Multilinguality.} \stanza{}'s architectural design is language-agnostic and data-driven, which allows us to release models supporting 66 languages, by training the pipeline on the Universal Dependencies (UD) treebanks and other multilingual corpora.
  \item \textbf{State-of-the-art performance.} We evaluate \stanza{} on a total of 112 datasets, and find its neural pipeline adapts well to text of different genres, achieving state-of-the-art or competitive performance at each step of the pipeline.
\end{itemize}

Additionally, \stanza{} features a Python interface to the widely used Java \corenlp{} package, allowing access to additional tools such as coreference resolution and relation extraction.

\stanza{} is fully open source and we make pretrained models for all supported languages and datasets available for public download.
We hope \stanza{} can facilitate multilingual NLP research and applications, and drive future research that produces insights from human languages.

%% file: architecture.tex
\section{System Design and Architecture}

At the top level, \stanza{} consists of two individual components: (1) a fully neural multilingual NLP pipeline; (2) a Python client interface to the Java Stanford CoreNLP software.
In this section we introduce their designs.

\subsection{Neural Multilingual NLP Pipeline} 

\stanza{}'s neural pipeline consists of models that range from tokenizing raw text to performing syntactic analysis on entire sentences (see \reffig{architecture}).
All components are designed with processing many human languages in mind, with high-level design choices capturing common phenomena in many languages and data-driven models that learn the difference between these languages from data.
Moreover, the implementation of \stanza{} components is highly modular, and reuses basic model architectures when possible for compactness.
We highlight the important design choices here, and refer the reader to \citet{qi2018universal} for modeling details.

\paragraph*{Tokenization and Sentence Splitting.}
When presented raw text, \stanza{} tokenizes it and groups tokens into sentences as the first step of processing.
Unlike most existing toolkits, \stanza{} combines tokenization and sentence segmentation from raw text into a single module.
This is modeled as a tagging problem over character sequences, where the model predicts whether a given character is the end of a token, end of a sentence, or end of a multi-word token (MWT, see \reffig{mwt}).%
\footnote{Following Universal Dependencies \cite{nivre2020universal}, we make a distinction between \emph{tokens} (contiguous spans of characters in the input text) and syntactic \emph{words}.
These are interchangeable aside from the cases of MWTs, where one token can correspond to multiple words.}
We choose to predict MWTs jointly with tokenization because this task is context-sensitive in some languages.

\begin{figure}
\small
    \centering
    \begin{tabular}{|p{7cm}|}
    \hline
        (fr) L'Association \textit{\uline{des}} Hôtels\\
        \textit{(en) The Association of Hotels}\\
        (fr) Il y a \textit{\uline{des}} hôtels en bas de la rue\\
        \textit{(en) There are hotels down the street}\\
    \hline
    \end{tabular}
    \caption{An example of multi-word tokens in French. The \textit{des} in the first sentence corresponds to two syntactic words, \textit{de} and \textit{les}; the second \textit{des} is a single word.}
    \label{fig:mwt}
\end{figure}

\paragraph*{Multi-word Token Expansion.}
Once MWTs are identified by the tokenizer, they are expanded into the underlying syntactic words as the basis of downstream processing.
This is achieved with an ensemble of a frequency lexicon and a neural sequence-to-sequence (seq2seq) model,
to ensure that frequently observed expansions in the training set are always robustly expanded while maintaining flexibility to model unseen words statistically.

\paragraph*{POS and Morphological Feature Tagging.} 
For each word in a sentence, \stanza{} assigns it a part-of-speech (POS), and analyzes its universal morphological features (UFeats, \eg, singular/plural, 1\textsuperscript{st}/2\textsuperscript{nd}/3\textsuperscript{rd} person, etc.).
To predict POS and UFeats, we adopt a bidirectional long short-term memory network (Bi-LSTM) as the basic architecture.
For consistency among universal POS (UPOS), treebank-specific POS (XPOS), and UFeats, we adopt the biaffine scoring mechanism from \citet{dozat2016deep} to condition XPOS and UFeats prediction on that of UPOS.

\paragraph*{Lemmatization.} 
\stanza{} also lemmatizes each word in a sentence to recover its canonical form (\eg, \emph{did}$\to$\emph{do}).
Similar to the multi-word token expander, \stanza{}'s lemmatizer is implemented as an ensemble of a dictionary-based lemmatizer and a neural seq2seq lemmatizer.
An additional classifier is built on the encoder output of the seq2seq model, to predict \emph{shortcuts} such as lowercasing and identity copy for robustness on long input sequences such as URLs.

\paragraph*{Dependency Parsing.} 
\stanza{} parses each sentence for its syntactic structure, where each word in the sentence is assigned a syntactic head that is either another word in the sentence, or in the case of the root word, an artificial root symbol.
We implement a Bi-LSTM-based deep biaffine neural dependency parser \cite{dozat2016deep}. 
We further augment this model with two linguistically motivated features: one that predicts the \emph{linearization} order of two words in a given language, and the other that predicts the typical distance in linear order between them.
We have previously shown that these features significantly improve parsing accuracy \cite{qi2018universal}.

\paragraph*{Named Entity Recognition.}
For each input sentence, \stanza{} also recognizes named entities in it (\eg, person names, organizations, etc.).
For NER we adopt the contextualized string representation-based sequence tagger from \citet{akbik2018contextual}.
We first train a forward and a backward character-level LSTM language model, and at tagging time we concatenate the representations at the end of each word position from both language models with word embeddings, and feed the result into a standard one-layer Bi-LSTM sequence tagger with a conditional random field (CRF)-based decoder.

\subsection{CoreNLP Client}
Stanford's Java CoreNLP software provides a comprehensive set of NLP tools especially for the English language. 
However, these tools are not easily accessible with Python, the programming language of choice for many NLP practitioners, due to the lack of official support.
To facilitate the use of CoreNLP from Python, we take advantage of the existing server interface in CoreNLP, and implement a robust client as its Python interface.

When the CoreNLP client is instantiated, \stanza{} will automatically start the CoreNLP server as a local process.
The client then communicates with the server through its RESTful APIs, after which annotations are transmitted in Protocol Buffers, and converted back to native Python objects.
Users can also specify JSON or XML as annotation format.
To ensure robustness, while the client is being used, \stanza{} periodically checks the health of the server, and restarts it if necessary.

%% file: usage.tex
\section{System Usage}

\stanza{}'s user interface is designed to allow quick out-of-the-box processing of multilingual text.
To achieve this, \stanza{} supports automated model download via Python code and pipeline customization with processors of choice.
Annotation results can be accessed as native Python objects to allow for flexible post-processing.

\subsection{Neural Pipeline Interface}

\stanza{}'s neural NLP pipeline can be initialized with the \texttt{Pipeline} class, taking language name as an argument.
By default, all processors will be loaded and run over the input text; however, users can also specify the processors to load and run with a list of processor names as an argument.
Users can additionally specify other processor-level properties, such as batch sizes used by processors, at initialization time.

The following code snippet shows a minimal usage of \stanza{} for downloading the Chinese model, annotating a sentence with customized processors, and printing out all annotations:
\begin{CJK*}{UTF8}{gbsn}
\begin{lstlisting}
import stanza
# download Chinese model
stanza.download('zh')
# initialize Chinese neural pipeline
nlp = stanza.Pipeline('zh', processors='tokenize,pos,ner')
# run annotation over a sentence
doc = nlp('%*\color{purple}\textrm{斯坦福是一所私立研究型大学。}*)')
print(doc)
\end{lstlisting}
\end{CJK*}

After all processors are run, a \texttt{Document} instance will be returned, which stores all annotation results.
Within a \texttt{Document}, annotations are further stored in \texttt{Sentence}s, \texttt{Token}s and \texttt{Word}s in a top-down fashion (\reffig{architecture}).
The following code snippet demonstrates how to access the text and POS tag of each word in a document and all named entities in the document:

\begin{lstlisting}
# print the text and POS of all words
for sentence in doc.sentences:
    for word in sentence.words:
        print(word.text, word.pos)

# print all entities in the document
print(doc.entities)
\end{lstlisting}

\stanza{} is designed to be run on different hardware devices.
By default, CUDA devices will be used whenever they are visible by the pipeline, or otherwise CPUs will be used.
However, users can force all computation to be run on CPUs by setting \texttt{use\_gpu=False} at initialization time.

\subsection{CoreNLP Client Interface}

The CoreNLP client interface is designed in a way that the actual communication with the backend CoreNLP server is transparent to the user.
To annotate an input text with the CoreNLP client, a \texttt{CoreNLPClient} instance needs to be initialized, with an optional list of CoreNLP annotators.
After the annotation is complete, results will be accessible as native Python objects. 

This code snippet shows how to establish a CoreNLP client and obtain the NER and coreference annotations of an English sentence:
\begin{lstlisting}
from stanza.server import CoreNLPClient

# start a CoreNLP client
with CoreNLPClient(annotators=['tokenize','ssplit','pos','lemma','ner','parse','coref']) as client:
    # run annotation over input
    ann = client.annotate('Emily said that she liked the movie.')
    # access all entities
    for sent in ann.sentence:
        print(sent.mentions)
    # access coreference annotations
    print(ann.corefChain)
\end{lstlisting}

With the client interface, users can annotate text in 6 languages as supported by CoreNLP.

\subsection{Interactive Web-based Demo}

\begin{figure}
    \centering
    \includegraphics[width=0.45\textwidth]{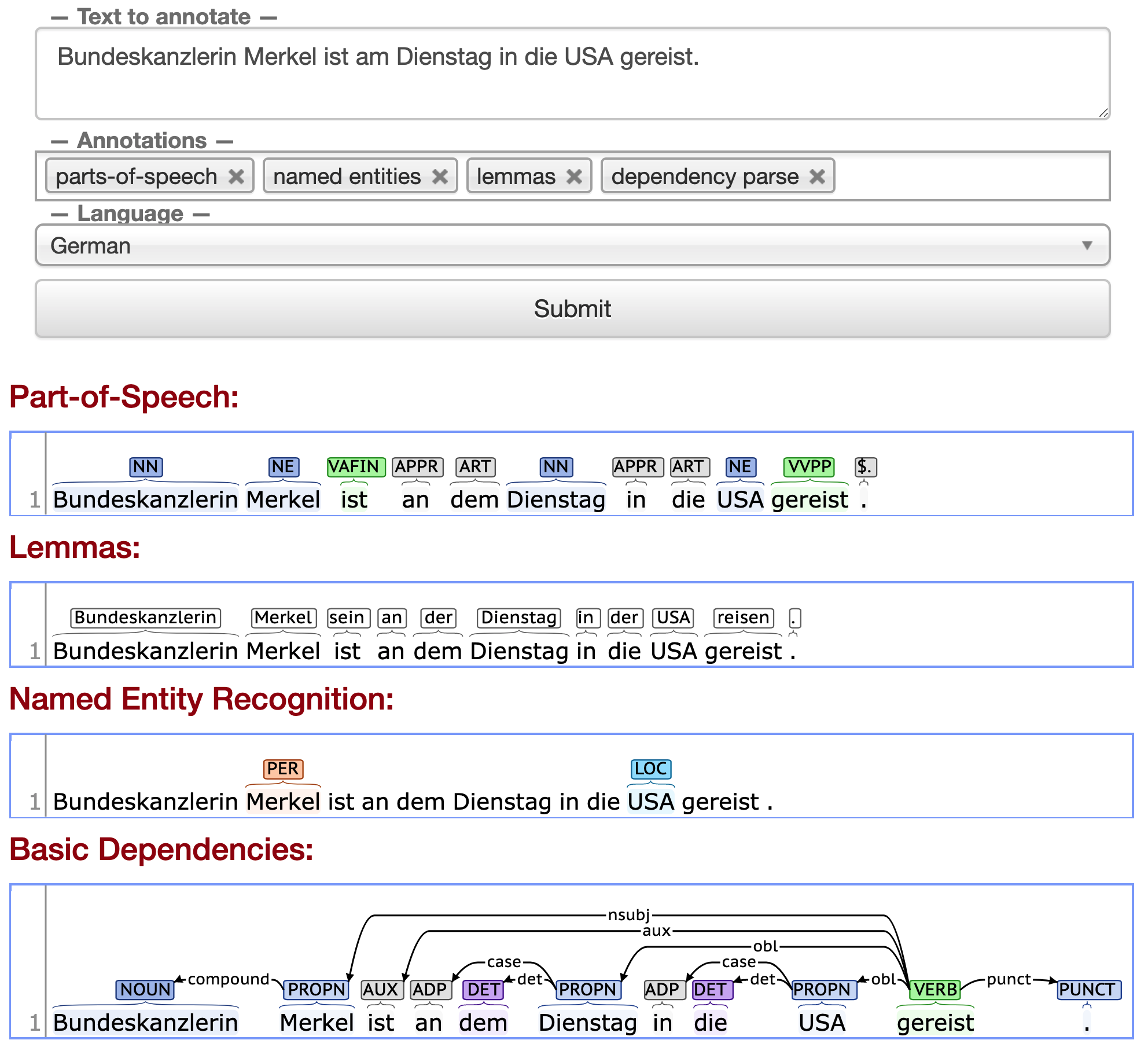}
    \caption{\stanza{} annotates a German sentence, as visualized by our interactive demo.
    Note \emph{am} is expanded into syntactic words \emph{an} and \emph{dem} before downstream analyses are performed.}
    \label{fig:interactive}
\end{figure}

To help visualize documents and their annotations generated by \stanza{}, we build an interactive web demo that runs the pipeline interactively.
For all languages and all annotations \stanza{} provides in those languages, we generate predictions from the models trained on the largest treebank/NER dataset, and visualize the result with the Brat rapid annotation tool.%
\footnote{\url{https://brat.nlplab.org/}}
This demo runs in a client\slash server architecture, and annotation is performed on the server side.
We make one instance of this demo publicly available at \url{http://stanza.run/}. 
It can also be run locally with proper Python libraries installed.
An example of running \stanza{} on a German sentence can be found in Figure \ref{fig:interactive}.

\begin{table*}[t]
\small
  \centering
  \renewcommand{\arraystretch}{1.2}
  \begin{tabular}{llcccccccccc}
    \toprule
     Treebank & System & Tokens & Sents. & Words & UPOS & XPOS & UFeats & Lemmas & UAS & LAS \\
     \midrule
     Overall (\ntreebank{} treebanks) & \stanzatitle & 99.09 & 86.05 & 98.63 & 92.49 & 91.80 & 89.93 & 92.78 & 80.45 & 75.68  \\
    \midrule
    \multirow{2}{3cm}{Arabic-PADT} & \stanzatitle & \textbf{99.98} & 80.43 & \textbf{97.88} & \textbf{94.89} & \textbf{91.75} & \textbf{91.86} & \textbf{93.27} & \textbf{83.27} & \textbf{79.33} \\
    & \udpipe & \textbf{99.98} & \textbf{82.09} & 94.58 & 90.36 & 84.00 & 84.16 & 88.46 & 72.67 & 68.14\\
    \midrule
    \multirow{2}{3cm}{Chinese-GSD} & \stanzatitle & \textbf{92.83} & 98.80 & \textbf{92.83} & \textbf{89.12} & \textbf{88.93} & \textbf{92.11} & \textbf{92.83} & \textbf{72.88} & \textbf{69.82} \\
    & \udpipe & 90.27 & \textbf{99.10} & 90.27 & 84.13 & 84.04 & 89.05 & 90.26 & 61.60 & 57.81 \\
    \midrule
    \multirow{3}{3cm}{English-EWT} & \stanzatitle & \textbf{99.01} & \textbf{81.13} & \textbf{99.01} & \textbf{95.40} & \textbf{95.12} & \textbf{96.11} & \textbf{97.21} & \textbf{86.22} & \textbf{83.59} \\
    & \udpipe & 98.90 & 77.40 & 98.90 & 93.26 & 92.75 & 94.23 & 95.45 & 80.22 & 77.03 \\
    & \spacy & 97.30 & 61.19 & 97.30 & 86.72 & 90.83 & -- & 87.05 & -- & --\\
    \midrule
    \multirow{3}{3cm}{French-GSD} & \stanzatitle & \textbf{99.68} & \textbf{94.92} & \textbf{99.48} & \textbf{97.30} & -- & \textbf{96.72} & \textbf{97.64} & \textbf{91.38} & \textbf{89.05}  \\
    & \udpipe & \textbf{99.68} & 93.59 & 98.81 & 95.85 & -- & 95.55 & 96.61 & 87.14 & 84.26 \\
    & \spacy & 98.34 & 77.30 & 94.15 & 86.82 & -- & -- & 87.29 & 67.46 & 60.60\\
    \midrule
    \multirow{3}{3cm}{Spanish-AnCora} & \stanzatitle & \textbf{99.98} & \textbf{99.07} & \textbf{99.98} & \textbf{98.78} & \textbf{98.67} & \textbf{98.59} & \textbf{99.19} & \textbf{92.21} & \textbf{90.01} \\
    & \udpipe & 99.97 & 98.32 & 99.95 & 98.32 & 98.13 & 98.13 & 98.48 & 88.22 & 85.10\\
    & \spacy & 99.47 & 97.59 & 98.95 & 94.04 & -- & -- & 79.63 & 86.63 & 84.13\\
    \bottomrule
  \end{tabular}
  \caption{Neural pipeline performance comparisons on the Universal Dependencies (v2.5) test treebanks.
  For our system we show macro-averaged results over all 100 treebanks.
  We also compare our system against \udpipe{} and \spacy{} on treebanks of five major languages where the corresponding pretrained models are publicly available.
  All results are \fone{} scores produced by the 2018 UD Shared Task official evaluation script.}
\label{tab:ud}
\end{table*}

\subsection{Training Pipeline Models}

For all neural processors, \stanza{} provides command-line interfaces for users to train their own customized models.
To do this, users need to prepare the training and development data in compatible formats (i.e., \texttt{CoNLL-U} format for the Universal Dependencies pipeline and \texttt{BIO} format column files for the NER model).
The following command trains a neural dependency parser with user-specified training and development data:
\begin{lstlisting}[language=bash]
$ python -m stanza.models.parser \
    --train_file train.conllu \
    --eval_file dev.conllu \
    --gold_file dev.conllu \
    --output_file output.conllu
\end{lstlisting}

%% file: experiments.tex
\section{Performance Evaluation}

To establish benchmark results and compare with other popular toolkits, we trained and evaluated \stanza{} on a total of \ndataset{} datasets.
All pretrained models are publicly downloadable.

\paragraph*{Datasets.}
We train and evaluate \stanza{}'s tokenizer/sentence splitter, MWT expander, POS/UFeats tagger, lemmatizer, and dependency parser with the Universal Dependencies v2.5 treebanks \cite{ud2.5}.
For training we use \ntreebank{} treebanks from this release that have non-copyrighted training data, and for treebanks that do not include development data, we randomly split out 20\% of the training data as development data.
These treebanks represent \nlang{} languages, mostly European languages, but spanning a diversity of language families, including Indo-European, Afro-Asiatic, Uralic, Turkic, Sino-Tibetan, etc.
For NER, we train and evaluate \stanza{} with \nnerdataset{} publicly available datasets covering \nnerlang{} major languages as shown in \reftab{ner} \cite{nothman2013learning, sang2003introduction, tjong2002introduction, benikova2014nosta, mohit2012recall, taule2008ancora, weischedel2013ontonotes}.
For the WikiNER corpora, as canonical splits are not available, we randomly split them into 70\% training, 15\% dev and 15\% test splits.
For all other corpora we used their canonical splits.

\paragraph*{Training.}
On the Universal Dependencies treebanks, we tuned all hyper-parameters on several large treebanks and applied them to all other treebanks.
We used the word2vec embeddings released as part of the 2018 UD Shared Task \cite{zeman2018conll}, or the fastText embeddings \cite{bojanowski2017enriching} whenever word2vec is not available.
For the character-level language models in the NER component, we pretrained them on a mix of the Common Crawl and Wikipedia dumps, and the news corpora released by the WMT19 Shared Task \cite{barrault2019findings}, except for English and Chinese, for which we pretrained on the Google One Billion Word \cite{chelba2013billion} and the Chinese Gigaword corpora%
\footnote{\url{https://catalog.ldc.upenn.edu/LDC2011T13}}, respectively.
We again applied the same hyper-parameters to models for all languages.

\paragraph*{Universal Dependencies Results.}
For performance on UD treebanks, we compared \stanza{} (v1.0) against \udpipe{} (v1.2) and \spacy{} (v2.2) on treebanks of 5 major languages whenever a pretrained model is available.
As shown in \reftab{ud}, \stanza{} achieved the best performance on most scores reported.
Notably, we find that \stanza{}'s language-agnostic architecture is able to adapt to datasets of different languages and genres.
This is also shown by \stanza{}'s high macro-averaged scores over \ntreebank{} treebanks covering \nlang{} languages.

\begin{table}[t]
  \small
  \centering
  \setlength{\tabcolsep}{0.3em}
  \newcommand{\pha}{\ph{$^\ast$}}
  \begin{tabular}{llcccc}
    \toprule
     Language & Corpus & \# Types & \stanzatitle & \flair & \spacy \\
    \midrule
    Arabic & AQMAR & 4 & \bf{74.3} & 74.0 & --\pha \\
    \midrule
    Chinese & OntoNotes & 18 & \bf{79.2} & -- & --\pha \\
    \midrule
    Dutch & CoNLL02 & 4 & 89.2 & \bf{90.3} & 73.8\pha \\
    & WikiNER & 4 & \bf{94.8} & \bf{94.8} & 90.9\pha \\
    \midrule
    English & CoNLL03 & 4 & 92.1 & \bf{92.7} & 81.0\pha \\
    & OntoNotes & 18 & 88.8 & \bf{89.0} & 85.4$^\ast$ \\
    \midrule
    French & WikiNER & 4 & \bf{92.9} & 92.5 & 88.8$^\ast$ \\
    \midrule
    German & CoNLL03 & 4 & 81.9 & \bf{82.5} & 63.9\pha \\
    & GermEval14 & 4 & 85.2 & \bf{85.4} & 68.4\pha \\
    \midrule
    Russian & WikiNER & 4 & \bf{92.9} & -- & --\pha \\
    \midrule
    Spanish & CoNLL02 & 4 & \bf{88.1} & 87.3 & 77.5\pha \\
    & AnCora & 4 & \bf{88.6} & 88.4 & 76.1\pha \\
    \bottomrule
  \end{tabular}
  \caption{NER performance across different languages and corpora. 
  All scores reported are entity micro-averaged test \fone{}.
  For each corpus we also list the number of entity types.
  $\ast$ marks results from publicly available pretrained models on the same dataset, while others are from models retrained on our datasets.}
\label{tab:ner}
\end{table}

\paragraph*{NER Results.}
For performance of the NER component, we compared \stanza{} (v1.0) against \flair{} (v0.4.5) and \spacy{} (v2.2).
For \spacy{} we reported results from its publicly available pretrained model whenever one trained on the same dataset can be found, otherwise we retrained its model on our datasets with default hyper-parameters, following the publicly available tutorial.%
\footnote{\url{https://spacy.io/usage/training##ner} Note that, following this public tutorial, we did not use pretrained word embeddings when training \spacy{} NER models, although using pretrained word embeddings may potentially improve the NER results.}
For \flair{}, since their downloadable models were pretrained on dataset versions different from canonical ones, we retrained all models on our own dataset splits with their best reported hyper-parameters.
All test results are shown in \reftab{ner}.
We find that on all datasets \stanza{} achieved either higher or close \fone{} scores when compared against \flair{}.
When compared to \spacy{}, \stanza{}'s NER performance is much better.
It is worth noting that \stanza{}'s high performance is achieved with much smaller models compared with \flair{} (up to 75\% smaller), as we intentionally compressed the models for memory efficiency and ease of distribution.

\paragraph{Speed comparison.} We compare \stanza{} against existing toolkits to evaluate the time it takes to annotate text (see \reftab{speed}). 
For GPU tests we use a single NVIDIA Titan RTX card.
Unsurprisingly, \stanza{}'s extensive use of accurate neural models makes it take significantly longer than \spacy{} to annotate text, but it is still competitive when compared against toolkits of similar accuracy, especially with the help of GPU acceleration.

%% file: conclusion.tex
\section{Conclusion and Future Work}

We introduced \stanza, a Python natural language processing toolkit supporting many human languages.
We have showed that \stanza{}'s neural pipeline not only has wide coverage of human languages, but also is accurate on all tasks, thanks to its language-agnostic, fully neural architectural design.
Simultaneously, \stanza{}'s CoreNLP client extends its functionality with additional NLP tools.

For future work, we consider the following areas of improvement in the near term:
\vspace{-0.4em}
\begin{itemize}
\setlength\itemsep{0em}
    \item Models downloadable in \stanza{} are largely trained on a single dataset. 
    To make models robust to many different genres of text, we would like to investigate the possibility of pooling various sources of compatible data to train ``default'' models for each language;
    \item The amount of computation and resources available to us is limited.
    We would therefore like to build an open ``model zoo'' for \stanza, so that researchers from outside our group can also contribute their models and benefit from models released by others;
    \item \stanza{} was designed to optimize for accuracy of its predictions, but this sometimes comes at the cost of computational efficiency and limits the toolkit's use.
    We would like to further investigate reducing model sizes and speeding up computation in the toolkit, while still maintaining the same level of accuracy.
    \item We would also like to expand \stanza{}'s functionality by adding other processors such as neural coreference resolution or relation extraction for richer text analytics.
\end{itemize}

\begin{table}
  \small
  \centering
  \setlength{\tabcolsep}{0.4em}
  \newcommand{\pha}{\ph{$^\ast$}}
  \begin{tabular}{lccccccc}
    \toprule
    \multirow{2}{1cm}{Task} & \multicolumn{2}{c}{\stanzatitle{}} && \udpipe{} && \multicolumn{2}{c}{\flair{}}\\
    \cline{2-3}\cline{5-5}\cline{7-8}
    & \textsc{cpu} & \textsc{gpu} && \textsc{cpu} && \textsc{cpu} & \textsc{gpu}\\
    \midrule 
    UD &  10.3$\times$ & 3.22$\times$ && 4.30$\times$ && -- & -- \\
    NER & 17.7$\times$ & 1.08$\times$ && -- && 51.8$\times$ & 1.17$\times$ \\
    \bottomrule
  \end{tabular}
  \caption{Annotation runtime of various toolkits relative to \spacy{} (CPU) on the English EWT treebank and OntoNotes NER test sets.
  For reference, on the compared UD and NER tasks, \spacy{} is able to process 8140 and 5912 tokens per second, respectively.}
\label{tab:speed}
\end{table}

%% file: acknowledgments.tex
\section*{Acknowledgments}

The authors would like to thank the anonymous reviewers for their comments, 
Arun Chaganty for his early contribution to this toolkit,
Tim Dozat for his design of the original architectures of the tagger and parser models,
Matthew Honnibal and Ines Montani for their help with \spacy{} integration and helpful comments on the draft,
Ranting Guo for the logo design,
and John Bauer and the community contributors for their help with maintaining and improving this toolkit.
This research is funded in part by Samsung Electronics Co., Ltd. and in part by the SAIL-JD Research Initiative.

\clearpage

%% file: main.bbl
\begin{thebibliography}{19}
\expandafter\ifx\csname natexlab\endcsname\relax\def\natexlab#1{#1}\fi

\bibitem[{Akbik et~al.(2019)Akbik, Bergmann, Blythe, Rasul, Schweter, and
  Vollgraf}]{akbik-etal-2019-flair}
Alan Akbik, Tanja Bergmann, Duncan Blythe, Kashif Rasul, Stefan Schweter, and
  Roland Vollgraf. 2019.
\newblock \href {https://www.aclweb.org/anthology/N19-4010} {{FLAIR}: An
  easy-to-use framework for state-of-the-art {NLP}}.
\newblock In \emph{Proceedings of the 2019 Conference of the North {A}merican
  Chapter of the Association for Computational Linguistics (Demonstrations)}.
  Association for Computational Linguistics.

\bibitem[{Akbik et~al.(2018)Akbik, Blythe, and Vollgraf}]{akbik2018contextual}
Alan Akbik, Duncan Blythe, and Roland Vollgraf. 2018.
\newblock \href {https://www.aclweb.org/anthology/C18-1139} {Contextual string
  embeddings for sequence labeling}.
\newblock In \emph{Proceedings of the 27th International Conference on
  Computational Linguistics}. Association for Computational Linguistics.

\bibitem[{Barrault et~al.(2019)Barrault, Bojar, Costa-juss{\`a}, Federmann,
  Fishel, Graham, Haddow, Huck, Koehn, Malmasi, Monz, M{\"u}ller, Pal, Post,
  and Zampieri}]{barrault2019findings}
Lo{\"\i}c Barrault, Ond{\v{r}}ej Bojar, Marta~R. Costa-juss{\`a}, Christian
  Federmann, Mark Fishel, Yvette Graham, Barry Haddow, Matthias Huck, Philipp
  Koehn, Shervin Malmasi, Christof Monz, Mathias M{\"u}ller, Santanu Pal, Matt
  Post, and Marcos Zampieri. 2019.
\newblock \href {https://www.aclweb.org/anthology/W19-5301} {Findings of the
  2019 conference on machine translation ({WMT}19)}.
\newblock In \emph{Proceedings of the Fourth Conference on Machine Translation
  (Volume 2: Shared Task Papers, Day 1)}. Association for Computational
  Linguistics.

\bibitem[{Benikova et~al.(2014)Benikova, Biemann, and
  Reznicek}]{benikova2014nosta}
Darina Benikova, Chris Biemann, and Marc Reznicek. 2014.
\newblock {NoSta-D} named entity annotation for {German}: Guidelines and
  dataset.
\newblock In \emph{Proceedings of the Ninth International Conference on
  Language Resources and Evaluation (LREC'14)}.

\bibitem[{Bojanowski et~al.(2017)Bojanowski, Grave, Joulin, and
  Mikolov}]{bojanowski2017enriching}
Piotr Bojanowski, Edouard Grave, Armand Joulin, and Tomas Mikolov. 2017.
\newblock \href {http://dx.doi.org/10.1162/tacl_a_00051} {Enriching word
  vectors with subword information}.
\newblock \emph{Transactions of the Association for Computational Linguistics},
  5.

\bibitem[{Chelba et~al.(2013)Chelba, Mikolov, Schuster, Ge, Brants, Koehn, and
  Robinson}]{chelba2013billion}
Ciprian Chelba, Tomas Mikolov, Mike Schuster, Qi~Ge, Thorsten Brants, Phillipp
  Koehn, and Tony Robinson. 2013.
\newblock \href {http://arxiv.org/abs/1312.3005} {One billion word benchmark
  for measuring progress in statistical language modeling}.
\newblock Technical report, Google.

\bibitem[{Dozat and Manning(2017)}]{dozat2016deep}
Timothy Dozat and Christopher~D. Manning. 2017.
\newblock \href {https://nlp.stanford.edu/pubs/dozat2017deep.pdf} {Deep
  biaffine attention for neural dependency parsing}.
\newblock In \emph{International Conference on Learning Representations
  (ICLR)}.

\bibitem[{Manning et~al.(2014)Manning, Surdeanu, Bauer, Finkel, Bethard, and
  McClosky}]{manning-EtAl:2014:P14-5}
Christopher~D. Manning, Mihai Surdeanu, John Bauer, Jenny Finkel, Steven~J.
  Bethard, and David McClosky. 2014.
\newblock \href {http://www.aclweb.org/anthology/P/P14/P14-5010} {The
  {Stanford} {CoreNLP} natural language processing toolkit}.
\newblock In \emph{Association for Computational Linguistics (ACL) System
  Demonstrations}.

\bibitem[{Mohit et~al.(2012)Mohit, Schneider, Bhowmick, Oflazer, and
  Smith}]{mohit2012recall}
Behrang Mohit, Nathan Schneider, Rishav Bhowmick, Kemal Oflazer, and Noah~A
  Smith. 2012.
\newblock Recall-oriented learning of named entities in {A}rabic {W}ikipedia.
\newblock In \emph{Proceedings of the 13th Conference of the European Chapter
  of the Association for Computational Linguistics}. Association for
  Computational Linguistics.

\bibitem[{Nivre et~al.(2020)Nivre, de~Marneffe, Ginter, Haji{\v c}, Manning,
  Pyysalo, Schuster, Tyers, and Zeman}]{nivre2020universal}
Joakim Nivre, Marie-Catherine de~Marneffe, Filip Ginter, Jan Haji{\v c},
  Christopher~D. Manning, Sampo Pyysalo, Sebastian Schuster, Francis Tyers, and
  Daniel Zeman. 2020.
\newblock Universal dependencies v2: An evergrowing multilingual treebank
  collection.
\newblock In \emph{Proceedings of the Twelfth International Conference on
  Language Resources and Evaluation (LREC'20)}.

\bibitem[{Nothman et~al.(2013)Nothman, Ringland, Radford, Murphy, and
  Curran}]{nothman2013learning}
Joel Nothman, Nicky Ringland, Will Radford, Tara Murphy, and James~R Curran.
  2013.
\newblock Learning multilingual named entity recognition from {W}ikipedia.
\newblock \emph{Artificial Intelligence}, 194:151--175.

\bibitem[{Qi et~al.(2018)Qi, Dozat, Zhang, and Manning}]{qi2018universal}
Peng Qi, Timothy Dozat, Yuhao Zhang, and Christopher~D. Manning. 2018.
\newblock \href {https://nlp.stanford.edu/pubs/qi2018universal.pdf} {Universal
  dependency parsing from scratch}.
\newblock In \emph{Proceedings of the {CoNLL} 2018 Shared Task: Multilingual
  Parsing from Raw Text to Universal Dependencies}. Association for
  Computational Linguistics.

\bibitem[{Straka(2018)}]{straka-2018-udpipe}
Milan Straka. 2018.
\newblock \href {https://www.aclweb.org/anthology/K18-2020} {{UDP}ipe 2.0
  prototype at {C}o{NLL} 2018 {UD} shared task}.
\newblock In \emph{Proceedings of the {C}o{NLL} 2018 Shared Task: Multilingual
  Parsing from Raw Text to Universal Dependencies}. Association for
  Computational Linguistics.

\bibitem[{Taul{\'e} et~al.(2008)Taul{\'e}, Mart{\'\i}, and
  Recasens}]{taule2008ancora}
Mariona Taul{\'e}, M.~Ant{\`o}nia Mart{\'\i}, and Marta Recasens. 2008.
\newblock \href
  {http://www.lrec-conf.org/proceedings/lrec2008/pdf/35_paper.pdf} {{A}n{C}ora:
  Multilevel annotated corpora for {C}atalan and {S}panish}.
\newblock In \emph{Proceedings of the Sixth International Conference on
  Language Resources and Evaluation ({LREC}'08)}. European Language Resources
  Association (ELRA).

\bibitem[{Tjong Kim~Sang(2002)}]{tjong2002introduction}
Erik~F. Tjong Kim~Sang. 2002.
\newblock \href {https://www.aclweb.org/anthology/W02-2024} {Introduction to
  the {C}o{NLL}-2002 shared task: Language-independent named entity
  recognition}.
\newblock In \emph{{COLING}-02: The 6th Conference on Natural Language Learning
  2002 ({C}o{NLL}-2002)}.

\bibitem[{Tjong Kim~Sang and De~Meulder(2003)}]{sang2003introduction}
Erik~F. Tjong Kim~Sang and Fien De~Meulder. 2003.
\newblock Introduction to the {CoNLL}-2003 shared task: Language-independent
  named entity recognition.
\newblock In \emph{Proceedings of the Seventh Conference on Natural Language
  Learning at HLT-NAACL 2003}.

\bibitem[{Weischedel et~al.(2013)Weischedel, Palmer, Marcus, Hovy, Pradhan,
  Ramshaw, Xue, Taylor, Kaufman, Franchini et~al.}]{weischedel2013ontonotes}
Ralph Weischedel, Martha Palmer, Mitchell Marcus, Eduard Hovy, Sameer Pradhan,
  Lance Ramshaw, Nianwen Xue, Ann Taylor, Jeff Kaufman, Michelle Franchini,
  et~al. 2013.
\newblock {OntoNotes} release 5.0.
\newblock \emph{Linguistic Data Consortium}.

\bibitem[{Zeman et~al.(2018)Zeman, Haji{\v{c}}, Popel, Potthast, Straka,
  Ginter, Nivre, and Petrov}]{zeman2018conll}
Daniel Zeman, Jan Haji{\v{c}}, Martin Popel, Martin Potthast, Milan Straka,
  Filip Ginter, Joakim Nivre, and Slav Petrov. 2018.
\newblock \href {https://www.aclweb.org/anthology/K18-2001} {{C}o{NLL} 2018
  shared task: Multilingual parsing from raw text to universal dependencies}.
\newblock In \emph{Proceedings of the {C}o{NLL} 2018 Shared Task: Multilingual
  Parsing from Raw Text to Universal Dependencies}. Association for
  Computational Linguistics.

\bibitem[{Zeman et~al.(2019)Zeman, Nivre, Abrams, Aepli, Agi{\'c}, Ahrenberg,
  Aleksandravi{\v c}i{\=u}t{\.e}, Antonsen, Aplonova, Aranzabe, Arutie,
  Asahara, Ateyah, Attia, Atutxa, Augustinus, Badmaeva, Ballesteros, Banerjee,
  Bank, Barbu~Mititelu, Basmov, Batchelor, Bauer, Bellato, Bengoetxea, Berzak,
  Bhat, Bhat, Biagetti, Bick, Bielinskien{\.e}, Blokland, Bobicev, Boizou,
  Borges~V{\"o}lker, B{\"o}rstell, Bosco, Bouma, Bowman, Boyd, Brokait{\.e},
  Burchardt, Candito, Caron, Caron, Cavalcanti, Cebiro{\u g}lu~Eryi{\u g}it,
  Cecchini, Celano, {\v C}{\'e}pl{\"o}, Cetin, Chalub, Choi, Cho, Chun,
  Cignarella, Cinkov{\'a}, Collomb, {\c C}{\"o}ltekin, Connor, Courtin,
  Davidson, de~Marneffe, de~Paiva, de~Souza, Diaz~de Ilarraza, Dickerson,
  Dione, Dirix, Dobrovoljc, Dozat, Droganova, Dwivedi, Eckhoff, Eli, Elkahky,
  Ephrem, Erina, Erjavec, Etienne, Evelyn, Farkas, Fernandez~Alcalde, Foster,
  Freitas, Fujita, Gajdo{\v s}ov{\'a}, Galbraith, Garcia, G{\"a}rdenfors,
  Garza, Gerdes, Ginter, Goenaga, Gojenola, G{\"o}k{\i}rmak, Goldberg,
  G{\'o}mez~Guinovart, Gonz{\'a}lez~Saavedra, Grici{\=u}t{\.e}, Grioni, Gr{\=
  u}z{\={\i}}tis, Guillaume, Guillot-Barbance, Habash, Haji{\v c}, Haji{\v
  c}~jr., H{\"a}m{\"a}l{\"a}inen, H{\`a}~M{\~y}, Han, Harris, Haug, Heinecke,
  Hennig, Hladk{\'a}, Hlav{\'a}{\v c}ov{\'a}, Hociung, Hohle, Hwang, Ikeda,
  Ion, Irimia, Ishola, Jel{\'{\i}}nek, Johannsen, J{\o}rgensen, Juutinen, Ka{\c
  s}{\i}kara, Kaasen, Kabaeva, Kahane, Kanayama, Kanerva, Katz, Kayadelen,
  Kenney, Kettnerov{\'a}, Kirchner, Klementieva, K{\"o}hn, Kopacewicz, Kotsyba,
  Kovalevskait{\.e}, Krek, Kwak, Laippala, Lambertino, Lam, Lando, Larasati,
  Lavrentiev, Lee, L{\^e}~H{\`{\^o}}ng, Lenci, Lertpradit, Leung, Li, Li, Li,
  Lim, Liovina, Li, Ljube{\v s}i{\'c}, Loginova, Lyashevskaya, Lynn, Macketanz,
  Makazhanov, Mandl, Manning, Manurung, M{\u a}r{\u a}nduc, Mare{\v c}ek,
  Marheinecke, Mart{\'{\i}}nez~Alonso, Martins, Ma{\v s}ek, Matsumoto,
  {McDonald}, {McGuinness}, Mendon{\c c}a, Miekka, Misirpashayeva, Missil{\"a},
  Mititelu, Mitrofan, Miyao, Montemagni, More, Moreno~Romero, Mori, Morioka,
  Mori, Moro, Mortensen, Moskalevskyi, Muischnek, Munro, Murawaki,
  M{\"u}{\"u}risep, Nainwani, Navarro~Hor{\~n}iacek, Nedoluzhko, Ne{\v
  s}pore-B{\=e}rzkalne, Nguy{\~{\^e}}n~Th{\d i}, Nguy{\~{\^e}}n Th{\d i}~Minh,
  Nikaido, Nikolaev, Nitisaroj, Nurmi, Ojala, Ojha, Ol{\'u}{\`o}kun, Omura,
  Osenova, {\"O}stling, {\O}vrelid, Partanen, Pascual, Passarotti, Patejuk,
  Paulino-Passos, Peljak-{\L}api{\'n}ska, Peng, Perez, Perrier, Petrova,
  Petrov, Phelan, Piitulainen, Pirinen, Pitler, Plank, Poibeau, Ponomareva,
  Popel, Pretkalni{\c n}a, Pr{\'e}vost, Prokopidis, Przepi{\'o}rkowski,
  Puolakainen, Pyysalo, Qi, R{\"a}{\"a}bis, Rademaker, Ramasamy, Rama, Ramisch,
  Ravishankar, Real, Reddy, Rehm, Riabov, Rie{\ss}ler, Rimkut{\.e}, Rinaldi,
  Rituma, Rocha, Romanenko, Rosa, Rovati, Rosca, Rudina, Rueter, Sadde, Sagot,
  Saleh, Salomoni, Samard{\v z}i{\'c}, Samson, Sanguinetti, S{\"a}rg,
  Saul{\={\i}}te, Sawanakunanon, Schneider, Schuster, Seddah, Seeker, Seraji,
  Shen, Shimada, Shirasu, Shohibussirri, Sichinava, Silveira, Silveira, Simi,
  Simionescu, Simk{\'o}, {\v S}imkov{\'a}, Simov, Smith, Soares-Bastos,
  Spadine, Stella, Straka, Strnadov{\'a}, Suhr, Sulubacak, Suzuki,
  Sz{\'a}nt{\'o}, Taji, Takahashi, Tamburini, Tanaka, Tellier, Thomas, Torga,
  Trosterud, Trukhina, Tsarfaty, Tyers, Uematsu, Ure{\v s}ov{\'a}, Uria,
  Uszkoreit, Utka, Vajjala, van Niekerk, van Noord, Varga, Villemonte de~la
  Clergerie, Vincze, Wallin, Walsh, Wang, Washington, Wendt, Williams,
  Wir{\'e}n, Wittern, Woldemariam, Wong, Wr{\'o}blewska, Yako, Yamazaki, Yan,
  Yasuoka, Yavrumyan, Yu, {\v Z}abokrtsk{\'y}, Zeldes, Zhang, and Zhu}]{ud2.5}
Daniel Zeman, Joakim Nivre, Mitchell Abrams, No{\"e}mi Aepli, {\v Z}eljko
  Agi{\'c}, Lars Ahrenberg, Gabriel{\.e} Aleksandravi{\v c}i{\=u}t{\.e}, Lene
  Antonsen, Katya Aplonova, Maria~Jesus Aranzabe, Gashaw Arutie, Masayuki
  Asahara, Luma Ateyah, Mohammed Attia, Aitziber Atutxa, Liesbeth Augustinus,
  Elena Badmaeva, Miguel Ballesteros, Esha Banerjee, Sebastian Bank, Verginica
  Barbu~Mititelu, Victoria Basmov, Colin Batchelor, John Bauer, Sandra Bellato,
  Kepa Bengoetxea, Yevgeni Berzak, Irshad~Ahmad Bhat, Riyaz~Ahmad Bhat, Erica
  Biagetti, Eckhard Bick, Agn{\.e} Bielinskien{\.e}, Rogier Blokland, Victoria
  Bobicev, Lo{\"{\i}}c Boizou, Emanuel Borges~V{\"o}lker, Carl B{\"o}rstell,
  Cristina Bosco, Gosse Bouma, Sam Bowman, Adriane Boyd, Kristina Brokait{\.e},
  Aljoscha Burchardt, Marie Candito, Bernard Caron, Gauthier Caron, Tatiana
  Cavalcanti, G{\"u}l{\c s}en Cebiro{\u g}lu~Eryi{\u g}it, Flavio~Massimiliano
  Cecchini, Giuseppe G.~A. Celano, Slavom{\'{\i}}r {\v C}{\'e}pl{\"o}, Savas
  Cetin, Fabricio Chalub, Jinho Choi, Yongseok Cho, Jayeol Chun, Alessandra~T.
  Cignarella, Silvie Cinkov{\'a}, Aur{\'e}lie Collomb, {\c C}a{\u g}r{\i} {\c
  C}{\"o}ltekin, Miriam Connor, Marine Courtin, Elizabeth Davidson,
  Marie-Catherine de~Marneffe, Valeria de~Paiva, Elvis de~Souza, Arantza
  Diaz~de Ilarraza, Carly Dickerson, Bamba Dione, Peter Dirix, Kaja Dobrovoljc,
  Timothy Dozat, Kira Droganova, Puneet Dwivedi, Hanne Eckhoff, Marhaba Eli,
  Ali Elkahky, Binyam Ephrem, Olga Erina, Toma{\v z} Erjavec, Aline Etienne,
  Wograine Evelyn, Rich{\'a}rd Farkas, Hector Fernandez~Alcalde, Jennifer
  Foster, Cl{\'a}udia Freitas, Kazunori Fujita, Katar{\'{\i}}na Gajdo{\v
  s}ov{\'a}, Daniel Galbraith, Marcos Garcia, Moa G{\"a}rdenfors, Sebastian
  Garza, Kim Gerdes, Filip Ginter, Iakes Goenaga, Koldo Gojenola, Memduh
  G{\"o}k{\i}rmak, Yoav Goldberg, Xavier G{\'o}mez~Guinovart, Berta
  Gonz{\'a}lez~Saavedra, Bernadeta Grici{\=u}t{\.e}, Matias Grioni, Normunds
  Gr{\= u}z{\={\i}}tis, Bruno Guillaume, C{\'e}line Guillot-Barbance, Nizar
  Habash, Jan Haji{\v c}, Jan Haji{\v c}~jr., Mika H{\"a}m{\"a}l{\"a}inen, Linh
  H{\`a}~M{\~y}, Na-Rae Han, Kim Harris, Dag Haug, Johannes Heinecke, Felix
  Hennig, Barbora Hladk{\'a}, Jaroslava Hlav{\'a}{\v c}ov{\'a}, Florinel
  Hociung, Petter Hohle, Jena Hwang, Takumi Ikeda, Radu Ion, Elena Irimia, {\d
  O}l{\'a}j{\'{\i}}d{\'e} Ishola, Tom{\'a}{\v s} Jel{\'{\i}}nek, Anders
  Johannsen, Fredrik J{\o}rgensen, Markus Juutinen, H{\"u}ner Ka{\c s}{\i}kara,
  Andre Kaasen, Nadezhda Kabaeva, Sylvain Kahane, Hiroshi Kanayama, Jenna
  Kanerva, Boris Katz, Tolga Kayadelen, Jessica Kenney, V{\'a}clava
  Kettnerov{\'a}, Jesse Kirchner, Elena Klementieva, Arne K{\"o}hn, Kamil
  Kopacewicz, Natalia Kotsyba, Jolanta Kovalevskait{\.e}, Simon Krek, Sookyoung
  Kwak, Veronika Laippala, Lorenzo Lambertino, Lucia Lam, Tatiana Lando,
  Septina~Dian Larasati, Alexei Lavrentiev, John Lee, Phương
  L{\^e}~H{\`{\^o}}ng, Alessandro Lenci, Saran Lertpradit, Herman Leung,
  Cheuk~Ying Li, Josie Li, Keying Li, {KyungTae} Lim, Maria Liovina, Yuan Li,
  Nikola Ljube{\v s}i{\'c}, Olga Loginova, Olga Lyashevskaya, Teresa Lynn,
  Vivien Macketanz, Aibek Makazhanov, Michael Mandl, Christopher Manning, Ruli
  Manurung, C{\u a}t{\u a}lina M{\u a}r{\u a}nduc, David Mare{\v c}ek, Katrin
  Marheinecke, H{\'e}ctor Mart{\'{\i}}nez~Alonso, Andr{\'e} Martins, Jan Ma{\v
  s}ek, Yuji Matsumoto, Ryan {McDonald}, Sarah {McGuinness}, Gustavo Mendon{\c
  c}a, Niko Miekka, Margarita Misirpashayeva, Anna Missil{\"a}, C{\u a}t{\u
  a}lin Mititelu, Maria Mitrofan, Yusuke Miyao, Simonetta Montemagni, Amir
  More, Laura Moreno~Romero, Keiko~Sophie Mori, Tomohiko Morioka, Shinsuke
  Mori, Shigeki Moro, Bjartur Mortensen, Bohdan Moskalevskyi, Kadri Muischnek,
  Robert Munro, Yugo Murawaki, Kaili M{\"u}{\"u}risep, Pinkey Nainwani,
  Juan~Ignacio Navarro~Hor{\~n}iacek, Anna Nedoluzhko, Gunta Ne{\v
  s}pore-B{\=e}rzkalne, Lương Nguy{\~{\^e}}n~Th{\d i}, Huy{\`{\^e}}n
  Nguy{\~{\^e}}n Th{\d i}~Minh, Yoshihiro Nikaido, Vitaly Nikolaev, Rattima
  Nitisaroj, Hanna Nurmi, Stina Ojala, Atul~Kr. Ojha, Ad{\'e}day{\d o}
  Ol{\'u}{\`o}kun, Mai Omura, Petya Osenova, Robert {\"O}stling, Lilja
  {\O}vrelid, Niko Partanen, Elena Pascual, Marco Passarotti, Agnieszka
  Patejuk, Guilherme Paulino-Passos, Angelika Peljak-{\L}api{\'n}ska, Siyao
  Peng, Cenel-Augusto Perez, Guy Perrier, Daria Petrova, Slav Petrov, Jason
  Phelan, Jussi Piitulainen, Tommi~A Pirinen, Emily Pitler, Barbara Plank,
  Thierry Poibeau, Larisa Ponomareva, Martin Popel, Lauma Pretkalni{\c n}a,
  Sophie Pr{\'e}vost, Prokopis Prokopidis, Adam Przepi{\'o}rkowski, Tiina
  Puolakainen, Sampo Pyysalo, Peng Qi, Andriela R{\"a}{\"a}bis, Alexandre
  Rademaker, Loganathan Ramasamy, Taraka Rama, Carlos Ramisch, Vinit
  Ravishankar, Livy Real, Siva Reddy, Georg Rehm, Ivan Riabov, Michael
  Rie{\ss}ler, Erika Rimkut{\.e}, Larissa Rinaldi, Laura Rituma, Luisa Rocha,
  Mykhailo Romanenko, Rudolf Rosa, Davide Rovati, Valentin Rosca, Olga Rudina,
  Jack Rueter, Shoval Sadde, Beno{\^{\i}}t Sagot, Shadi Saleh, Alessio
  Salomoni, Tanja Samard{\v z}i{\'c}, Stephanie Samson, Manuela Sanguinetti,
  Dage S{\"a}rg, Baiba Saul{\={\i}}te, Yanin Sawanakunanon, Nathan Schneider,
  Sebastian Schuster, Djam{\'e} Seddah, Wolfgang Seeker, Mojgan Seraji,
  Mo~Shen, Atsuko Shimada, Hiroyuki Shirasu, Muh Shohibussirri, Dmitry
  Sichinava, Aline Silveira, Natalia Silveira, Maria Simi, Radu Simionescu,
  Katalin Simk{\'o}, M{\'a}ria {\v S}imkov{\'a}, Kiril Simov, Aaron Smith,
  Isabela Soares-Bastos, Carolyn Spadine, Antonio Stella, Milan Straka, Jana
  Strnadov{\'a}, Alane Suhr, Umut Sulubacak, Shingo Suzuki, Zsolt
  Sz{\'a}nt{\'o}, Dima Taji, Yuta Takahashi, Fabio Tamburini, Takaaki Tanaka,
  Isabelle Tellier, Guillaume Thomas, Liisi Torga, Trond Trosterud, Anna
  Trukhina, Reut Tsarfaty, Francis Tyers, Sumire Uematsu, Zde{\v n}ka Ure{\v
  s}ov{\'a}, Larraitz Uria, Hans Uszkoreit, Andrius Utka, Sowmya Vajjala,
  Daniel van Niekerk, Gertjan van Noord, Viktor Varga, Eric Villemonte de~la
  Clergerie, Veronika Vincze, Lars Wallin, Abigail Walsh, Jing~Xian Wang,
  Jonathan~North Washington, Maximilan Wendt, Seyi Williams, Mats Wir{\'e}n,
  Christian Wittern, Tsegay Woldemariam, Tak-sum Wong, Alina Wr{\'o}blewska,
  Mary Yako, Naoki Yamazaki, Chunxiao Yan, Koichi Yasuoka, Marat~M. Yavrumyan,
  Zhuoran Yu, Zden{\v e}k {\v Z}abokrtsk{\'y}, Amir Zeldes, Manying Zhang, and
  Hanzhi Zhu. 2019.
\newblock \href {http://hdl.handle.net/11234/1-3105} {Universal {D}ependencies
  2.5}.
\newblock {LINDAT}/{CLARIAH}-{CZ} digital library at the Institute of Formal
  and Applied Linguistics ({{\'U}FAL}), Faculty of Mathematics and Physics,
  Charles University.

\end{thebibliography}
